# Bayesian Neural Word Embedding

**Oren Barkan**

Tel Aviv University, Israel
Microsoft, Israel

**Abstract**

Recently, several works in the domain of natural language processing presented successful methods for word embedding. Among them, the Skip-Gram with negative sampling, known also as word2vec, advanced the state-of-the-art of various linguistics tasks. In this paper, we propose a scalable Bayesian neural word embedding algorithm. The algorithm relies on a Variational Bayes solution for the Skip-Gram objective and a detailed step by step description is provided. We present experimental results that demonstrate the performance of the proposed algorithm for word analogy and similarity tasks on six different datasets and show it is competitive with the original Skip-Gram method.

## 1 Introduction and Related Work

Recent progress in neural word embedding methods has advanced the state-of-the-art in the domain of natural language processing (NLP) (Pennington, Socher, and Manning 2014; Collobert and Weston 2008; Mnih and Hinton 2008; Mikolov et al. 2013; Vilnis and Mccallum 2015; Zhang et al. 2014). These methods attempt to learn a low dimensional representation for words that captures semantic and syntactic relations. Specifically, Skip-Gram (SG) with negative sampling, known also as word2vec (Mikolov et al. 2013), set new records in various linguistic tasks and its applications have been extended to other domains beyond NLP such as computer vision (Frome, Corrado, and Shlens 2013; Lazaridou, Nghia, and Baroni 2015) and Collaborative Filtering (CF) (Barkan and Koenigstein 2016; Barkan, Brumer, and Koenigstein 2016).

In this paper, we propose a scalable Bayesian neural word embedding algorithm that in principle, can be applied to any dataset that is given as sets / sequences of items. Hence, the proposed method is not limited to the task of word embedding and may be applicable to general item similarity tasks as well. We provide a fully detailed step by step algorithm, which is straightforward to implement and requires negligible amount of parameter tuning.

Bayesian methods for words representation are proposed in (Vilnis and Mccallum 2015; Zhang et al. 2014). Different from these methods, we propose a Variational Bayes (VB) (Bishop 2006) solution to the SG objective. Therefore, we name our method Bayesian Skip-Gram (BSG).

VB solutions provides a stable and robust behavior that require negligible effort in hyperparameter tuning (Bishop 2006). This is in contrast to point estimate solutions that are more sensitive to singularities in the objective and often require significant amount of hyperparameter tuning.

While the SG method maps words to vectors, BSG maps words to densities in a latent space. Moreover, BSG provides for a confidence level in the embedding and opens up for density based similarities measures.

Our contribution is twofold: first, we derive a tractable (yet scalable) Bayesian solution to the SG objective and provide a detailed step by step algorithm. Secondly, we propose several density based similarity measures that can be investigated in further research.

The rest of the paper is organized as follows: Section 2 overviews the SG method. In Section 3, we provide the mathematical derivation of the BSG solution. In Section 4, we describe the BSG algorithm in detail. In Section 5, we present evaluations on six different datasets, where we show that in most cases BSG outperforms SG.

## 2 Skip-Gram with Negative Sampling

SG with negative sampling is a neural word embedding method that was introduced in (Mikolov et al. 2013). The method aims at estimating words representation that captures semantic and syntactic relations between a word to its surrounding words in a sentence. Note that SG can be applied with hierarchical softmax (Mikolov et al. 2013), but in this work we refer to SG as SG with negative sampling. The rest of this section provides a brief overview of the SG method.

Given a sequence of words $(w_i)_{i=1}^{L}$ from a finite vo-

cabulary $W = \{w_i\}_{i=1}^l$, SG aims at maximizing the following objective

$$\frac{1}{L}\sum_{i=1}^{L}\sum_{-c \leq j \leq c, j \neq 0} \log p(w_{i+j} | w_i) \qquad (1)$$

where $c$ is defined as the context window size and $p(w_j | w_i)$ is the softmax function:

$$p(w_j | w_i) = \frac{\exp(u_i^T v_j)}{\sum_{k \in I_W} \exp(u_i^T v_k)} \qquad (2)$$

$u_i \in U(\subset \mathbb{R}^m)$ and $v_i \in V(\subset \mathbb{R}^m)$ are latent vectors that correspond to the target and context representations for the word $w_i \in W$, respectively. $I_W \triangleq \{1,...,l\}$ and the parameter $m$ is chosen empirically and according to the size of the dataset.

Using Eq. (2) is impractical due to the computational complexity of $\nabla p(w_j | w_i)$, which is linear in $l$ that is usually in size of $10^5 - 10^6$.

## 2.1 Negative Sampling

Negative sampling (Mikolov et al. 2013) is introduced in order to overcome the above computational problem by the replacement of the softmax function from Eq. (2) with

$$p(w_j | w_i) = \sigma(u_i^T v_j) \prod_{k=1}^{N} \sigma(-u_i^T v_k) \qquad (3)$$

where $\sigma(x) = 1/1 + \exp(-x)$, $N$ is a parameter that determines the number of negative examples to be sampled per a positive example. A negative word $w_k$ is sampled from the unigram distribution raised to the 3/4rd power $p_{uni}^{3/4}(w)$, where $p_{uni}(w)$ is defined as the number of times $w$ appear in the entire corpus divided by the total length (in words) of the corpus. This distribution was found to outperform both the uniform and unigram distributions (Mikolov et al. 2013). The latent representations $U$ and $V$ are estimated by applying a stochastic gradient ascent with respect to the objective in Eq. (1).

It is worth noting that some versions of word embedding algorithms incorporate bias terms into Eq. (3) as follows

$$p(w_j | w_i) = \sigma(u_i^T v_j + b_i + b_j) \prod_{k=1}^{N} \sigma(-u_i^T v_k - b_i - b_k).$$

These biases often explain properties such as frequency of a word in the text (Pennington, Socher, and Manning 2014; Mnih and Hinton 2008) or popularity of an item in the dataset (Paquet and Koenigstein 2013). In this work, we do not use biases, since in our initial experiments their contribution was found to be marginal.

## 2.2 Data Subsampling

In order to overcome the imbalance between rare and frequent words the following subsampling procedure is suggested: Given the input words sequence, we discard each word $w$ in a sentence with a probability $p(discard | w) = 1 - \sqrt{\rho / f(w)}$ where $f(w)$ is the frequency of the word $w$ and $\rho$ is a prescribed threshold. This technique is reported to accelerate the learning process and improve the representation of rare words (Mikolov et al. 2013).

## 2.3 Word Representation and Similarity

SG produces two different representations $u_i$ and $v_i$ for the word $w_i$. In this work, we use $u_i$ as the final representation for $w_i$. Alternatively, we can use $v_i$, the additive composition $u_i + v_i$ or the concatenation $\begin{bmatrix} u_i^T & v_i^T \end{bmatrix}^T$. The last two options are reported to procure superior representation (Garten et al. 2015). The similarity between a pair of words $w_a, w_b$ is computed by applying the cosine similarity to their corresponding representations $u_a, u_b$ as follows

$$sim(w_a, w_b) = \frac{u_a^T u_b}{\|u_a\| \|u_b\|}.$$

## 3 Bayesian Skip-Gram (BSG)

As described in Section 2, SG produces point estimates for $U$ and $V$. In this section, we propose a method for deriving full distributions for $U$ and $V$ (in this paper, we use the terms *distribution* and *density* interchangeably). We assume that each target and context random vectors are independent and have normal priors with a zero mean and diagonal covariance with a precision parameter $\tau$ as follows

$$\forall i \in I_W: \ p(u_i) = N_{u_i}(0, \tau^{-1}I) \text{ and } p(v_i) = N_{v_i}(0, \tau^{-1}I).$$

Note that different values of $\tau$ can be set to the priors over $U$ and $V$. Furthermore, these hyperparameters can be treated as random variables and be learned from the data (Paquet and Koenigstein 2013). However, in this work, we assume these hyperparameters are given and identical.

We define $C(w_i)$ as a multiset that contains the indices of the context words of $w_i$ in the corpus. Let $I_P = \{(i, j) | j \in C(w_i)\}$ and $I_N = \{(i, j) | j \notin C(w_i)\}$ be the positive and negative sets, respectively. Note that $I_P$ is a multiset too and $I_N$'s size might be quadratic in the vocabulary size $l$. Therefore, we approximate $I_N$ by negative sampling as described in Section 2.1.

Let $I_D = I_P \cup I_N$ and define $D = \{d_{ij} \mid (i,j) \in I_D\}$, where $d : I_D \to \{1, -1\}$ is a random variable

$$d_{ij} \triangleq d((i,j)) = \begin{cases} 1 & (i,j) \in I_p \\ -1 & (i,j) \in I_N \end{cases}.$$

Then, the likelihood of $d_{ij}$ is given by $p(d_{ij} \mid u_i, v_j) = \sigma(d_{ij} u_i^T v_j)$. Note that when applied to multisets, the operator $\cup$ is defined as the multiset sum and not as the multiset union.

The joint distribution of $U, V$ and $D$ is given by

$$\begin{aligned} p(U,V,D) &= p(D \mid U,V) p(U) p(V) = \\ &\prod_{(i,j) \in I_D} p(d_{ij} \mid u_i, v_j) \prod_{i \in I_W} p(u_i) \prod_{i \in I_W} p(v_i) = \\ &\prod_{(i,j) \in I_D} \sigma(d_{ij} u_i^T v_j) \prod_{i \in I_W} N_{u_i}(0, \tau^{-1} I) \prod_{i \in I_W} N_{v_i}(0, \tau^{-1} I). \end{aligned} \quad (4)$$

## 3.1 Variational Approximation

We aim at computing the posterior $p(U,V \mid D)$. However, a direct computation is hard. The posterior can be approximated using MCMC approaches such as Gibbs sampling or by using VB methods. In this work, we choose to apply VB approximation (Bishop 2006), since it was shown to converge faster to an accurate solution, empirically.

Let $\theta = U \cup V$, VB approximates the posterior $P(\theta \mid D)$ by finding a fully factorized distribution

$$q(\theta) = q(U,V) = q(U)q(V) = \prod_{i=1}^{l} q(u_i) q(v_i)$$

that minimizes the KL divergence (Bishop 2006)

$$D_{KL}\left(q(\theta) \parallel P(\theta \mid D)\right) = \int q(\theta) \log \frac{q(\theta)}{p(\theta \mid D)} d\theta.$$

To this end, we define the following expression:

$$\begin{aligned} L(q) &\triangleq \int q(\theta) \log \frac{p(\theta, D)}{q(\theta)} d\theta = \\ &\int q(\theta) \log \frac{p(\theta \mid D)}{q(\theta)} d\theta + \int q(\theta) \log p(D) d\theta = \\ &-D_{KL}\left(q(\theta) \parallel p(\theta \mid D)\right) + \log p(D) \end{aligned} \quad (5)$$

where the last transition in Eq. (5) is due to the fact $q$ is a PDF. By rearranging Eq. (5) we get the relation $D_{KL}(q(\theta) \parallel p(\theta \mid D)) = \log p(D) - L(q)$, where we notice that $\log p(D)$ is independent of $q$. Hence, minimizing $D_{KL}(q(\theta) \parallel p(\theta \mid D))$ is equivalent to maximizing $L(q)$. It was shown (Bishop 2006) that $L(q)$ is maximized by an iterative procedure that is guaranteed to converge to a local optimum. This is done by updating each of $q$'s factors, sequentially and according to the update rule

$$q_{u_i}^* = \exp\left(\mathbf{E}_{q_{\theta \setminus u_i}}[\log p(\theta, D)] + const\right) \quad (6)$$

where the update for $q_{v_i}^*$ is performed by the replacement of $u_i$ with $v_i$ in Eq. (6).

Recall that the term $p(\theta, D) = p(U,V,D)$ in Eq. (6) contains the likelihood $p(D \mid U,V)$ from Eq. (4), which is a product of sigmoid functions of $U$ and $V$. Therefore, a conjugate relation between the likelihood and the priors does not hold and the distribution that is implied by $\mathbf{E}_{q_{\theta \setminus u_i}}[\log p(\theta, D)]$ in Eq. (6) does not belong to the exponential family.

Next, we show that by the introduction of an additional parameter $\xi_{ij}$ we are able to bring Eq. (6) to a form that is recognized as the Gaussian distribution.

We start by lower bounding $p(D \mid U,V)$ using the following logistic bound (Jaakkola and Jordan 1996):

$$\log \sigma(a) \geq \frac{a - \xi}{2} - \lambda(\xi)(a^2 - \xi^2) + \log \sigma(\xi) \quad (7)$$

where $\lambda(\xi) = \frac{1}{2\xi}\left(\sigma(\xi) - \frac{1}{2}\right)$. By applying the lower bound from (7) to $\log p(D \mid \theta)$ we get

$$\begin{aligned} &\log p(D \mid \theta) \geq \log p_\xi(D \mid \theta) = \\ &\sum_{(i,j) \in I_D} \frac{d_{ij} u_i^T v_j - \xi_{ij}}{2} - \lambda(\xi_{ij})(u_i^T v_j v_j^T u_i - \xi_{ij}^2) + \log \sigma(\xi_{ij}). \end{aligned} \quad (8)$$

By using Eq. (8) we bound $L(q)$ as follows

$$\begin{aligned} L(q) &\geq L_\xi(q) = \int q(\theta) \log \frac{p_\xi(\theta, D)}{q(\theta)} d\theta = \\ &\int q(\theta) \log \frac{p_\xi(D \mid \theta) p(\theta)}{q(\theta)} d\theta. \end{aligned}$$

Furthermore, it was shown (Jaakkola and Jordan 1996) that the bound in Eq. (6) is tight when

$$\begin{aligned} \xi_{ij}^2 &= \mathbf{E}_q[u_i^T v_j v_j^T u_i] = \text{var}(u_i^T v_j) + \mathbf{E}_q[u_i^T v_j]\mathbf{E}_q[v_j^T u_i] = \\ &\text{var}(u_i^T v_j) + \mu_{u_i}^T \mu_{v_j} \mu_{v_j}^T \mu_{u_i} \end{aligned} \quad (9)$$

where $\mu_{v_j} \triangleq \mathbf{E}_q[v_j]$ and the last transition in Eq. (9)

holds since $u_i$ and $v_j$ are independent. By assuming diagonal covariance matrices, the term $\text{var}(u_i^T v_j)$ in Eq. (8) is computed by

$$\text{var}(u_i^T v_j) = \text{var}\left(\sum_{k=1}^m u_{ik} v_{jk}\right) = \sum_{k=1}^m \text{var}(u_{ik} v_{jk}) =$$
$$\sum_{k=1}^m \sigma_{u_{ik}}^2 \sigma_{v_{jk}}^2 + \sigma_{u_{ik}}^2 \mu_{v_{jk}}^2 + \sigma_{v_{jk}}^2 \mu_{u_{ik}}^2 \quad . \quad (10)$$

Finally, by combining Eqs. (9) and (10) we get

$$\xi_{ij} = \sqrt{\sum_{k=1}^m (\sigma_{u_{ik}}^2 + \mu_{u_{ik}}^2)(\sigma_{v_{jk}}^2 + \mu_{v_{jk}}^2)} \quad . \quad (11)$$

Therefore, instead of maximizing $L(q)$ we can maximize $L_\xi(q)$ and this is done by replacing the term $\log p(\theta, D)$ from Eq. (6) with $\log p_\xi(\theta, D)$ as follows

$$q_{u_i}^* = \exp\left(\mathbf{E}_{q_{\theta\setminus u_i}}[\log p_\xi(\theta, D)] + const\right). \quad (12)$$

By applying the natural logarithm to Eq. (12) we get

$$\log q_{u_i}^* = \mathbf{E}_{q_{\theta\setminus u_i}}[\log p_\xi(\theta, D)] + const =$$
$$\mathbf{E}_{q_{\theta\setminus u_i}}[\log p_\xi(D|U,V)] + \mathbf{E}_{q_{\theta\setminus u_i}}[\log p(U,V)] + const = \quad (13)$$
$$u_i^T r_{u_i} - \frac{1}{2} u_i^T P_{u_i} u_i + const$$

where

$$P_{u_i} = \sum_{j \in I_{u_i}} \left[2\lambda(\xi_{ij}) \mathbf{E}_q[v_j v_j^T]\right] + \tau I$$
$$r_{u_i} = \frac{1}{2} \sum_{j \in I_{u_i}} d_{ij} \mu_{v_j} \quad (14)$$

with $I_{u_i} = \{j \mid (i,j) \in I_D\}$ and $\mathbf{E}_q[v_j v_j^T] = \Sigma_{v_j} + \mu_{v_j} \mu_{v_j}^T$. Note that in the last transition in Eq. (13), all terms that are independent of $u_i$ are absorbed into the *const* term.

By inspecting Eqs. (13) and (14), we see that $q_{u_i}^*$ is normally distributed with the natural parameters $P_{u_i} = \Sigma_{u_i}^{-1}$ (the precision matrix) and $r_{u_i} = P_{u_i} \mu_{u_i}$ (the means times precision vector). Note that the computation of $q_{v_i}^*$'s parameters is symmetric. Moreover, since the updates for $\{q_{u_i}\}_{i=1}^l$ depend only on $\{q_{v_i}\}_{i=1}^l$ and vice versa, they can be performed in parallel. This gives an alternating updates scheme that is embarrassingly parallel and (under the assumption of constant dataset) guaranteed to converge to a local optimum (Bishop 2006): First, update all $\{q_{u_i}\}_{i=1}^l$ (in parallel), then update all $\{q_{v_i}\}_{i=1}^l$ (in parallel) and repeat until convergence.

## 3.2 Stochastic Updates

Due to data sampling, the effective dataset changes between the iterations and the optimization becomes stochastic. Since we do not want to ignore the information from previous steps, we need to figure out a way to consider this information in our updates. A common practice is to apply updates in the spirit of the Robbins-Monro method (Robbins and Monro 1951). This is performed by the introduction of an iteration dependent variable $\beta^{(k)}$ that controls the updates as follows

$$P_{u_i}^{(k)} = \beta^{(k)} P_{u_i} + (1 - \beta^{(k)}) P_{u_i}^{(k-1)}$$
$$r_{u_i}^{(k)} = \beta^{(k)} r_{u_i} + (1 - \beta^{(k)}) r_{u_i}^{(k-1)} \quad .$$

In practice, this means that during the runtime of the algorithm we need to keep the results from the previous iteration.

Robbins and Monro showed several conditions for convergence, where one of them states that $\beta^{(k)}$ needs to satisfy:

$$\sum_{k=0}^\infty \beta^{(k)} = \infty \quad \text{and} \quad \sum_{k=0}^\infty (\beta^{(k)})^2 < \infty \quad . \quad (15)$$

To this end, we suggest to use $\beta^{(k)} = k^{-\gamma}$ with a decay parameter $0.5 < \gamma \leq 1$ as this ensures the conditions in (15) hold. We further suggest to set $\beta^{(k)} = 1$ for the first few iterations, in order to avoid too early convergence. Specifically, in our implementation, we did not perform stochastic updates in the first 10 iterations.

## 4 The BSG Algorithm

In this section, we provide a detailed description of the BSG algorithm that is based on Sections 2 and 3. The algorithm is described in Fig. 1 and includes three main stages. The first stage is an initialization, then the algorithm iterates between data sampling and parameter updates till a convergence criterion is met or number of iterations is exceeded. In what follows, we explain each of these stages in detail.

### 4.1 Stage 1 - Initialization

The algorithm is given the following hyperparameters: the input text $T$ (set of senstences), target representation dimension $m$, the number of maximum iterations $K$, the maximal window size $c_{\max}$, the negative to positive ratio $N \in \mathbb{N}$, the subsampling parameter $\rho$, a stopping threshold $\varepsilon$, the decay parameter $\gamma$ and the prior precision parameter $\tau$. As described in Section 3, different values of $\tau$ can be learned for $U$ and $V$, however in our implementation, we chose to use a unique parameter $\tau$.

In this stage, we further need to determine the effective set of words $W$ to learn representation for. This can be done by considering all words in the data that appear more than a prescribed number of times, or by considering the $l$ most popular words. In this work, we stick with the latter. Then, every word $w \notin W$ is discarded from the data (step 1).

Step 2 initializes the target distributions $Q = \{q_{u_i}, q_{v_i}\}_{i=1}^{l}$ parameters. Specifically, the means are drawn from the multivariate standard normal distribution and the covariance matrices are set to identity.

In step 3, we compute $p_{uni}$ according to the description in Section 2.1, then we raise it to the ¾ rd power.

Step 4 updates $k$ and $K$ according to $\kappa$. This ensures the stochastic updates are not performed in the first $\kappa$ iterations.

### 4.2 Stage 2 – Data Sampling

At the beginning of every iteration, we subsample the data (step 5.1) as described in Section 2.2. Then, we follow the description in Section 3: for each instance of the word $w_i$ in $T$, we sample a window size $c$ from the uniform discrete distribution over the set $\{1,...,c_{max}\}$ and consider the $c$ words to the left and to the right of $w_i$ as context words for $w_i$. This results in a multiset $C(w_i)$ that contains the indices of the context words of $w_i$ (an index may appear multiple times). Then, we create a positive multiset of tuples $I_P = \{(i,j) \mid j \in C(w_i)\}$ (step 5.3).

Next, for each tuple $(i,j) \in I_P$ we sample $N$ negative examples $(i,z)$ such that $(i,z) \notin I_P$. A negative word $w_z$ is sampled according to $p_{uni}^{3/4}(w_z)$. We further update $I_{u_i}, I_{v_j}, I_{v_z}, d_{ij}, d_{iz}$ accordingly (step 5.4).

An alternative implementation of step 5.4 is to save for each tuple a counter for the number of times it appears. This can be done by maintaining dictionary data structures that count positive and negative examples. This avoids the need of maintaining $\{I_{u_i}, I_{v_i}\}_{i=1}^{l}$ as multisets (list data structures) and replace them with set data structures.

### 4.3 Stage 3 – Parameter Updates

In this stage, we update the parameters of the distributions $Q = \{q_{u_i}, q_{v_i}\}_{i=1}^{l}$. The updates are performed first for $\{q_{u_i}\}_{i=1}^{l}$ (step 5.6) and then for $\{q_{v_i}\}_{i=1}^{l}$ in a symmetric manner (step 5.7). Moreover, each sequence of updates is performed in parallel. Note that step 5.6.1 involves the computation of $\xi_{ij}$, $\lambda(\xi_{ij})$ and $\mathbf{E}_q[v_j v_j^T]$ that are given by Eqs. (11), (7) and (14), respectively.

Due to data sampling the dataset is changed per iteration. Therefore, we apply stochastic updates (step 5.6.2). The stochastic updates are performed starting from the iteration $\kappa+1$. This is ensured by step 5.5.

A crucial point to notice is the computation of the mean and the covariance: first, we compute the covariance matrix by the inversion of the precision matrix (step 5.6.3). This is performed by using Cholesky decomposition. Then, we extract the mean (step 5.6.4). Finally, we set all the off diagonal values in $\Sigma_{u_i}$ to zeros (step 5.6.5) while keeping $P_{u_i}$ as is.

The algorithm stops if the convergence criterion is met or number of iterations is exceeded (last line).

### 4.4 Similarity Measures

BSG maps words to normal distributions. In this work, we choose to use the distributions $\{q_{u_i}\}_{i=1}^{l}$ for representing words. The similarity between a pair of words $w_i, w_j$ can be computed by the cosine similarity of their means $\mu_{u_i}, \mu_{u_j}$. By using the covariance, a confidence level can be computed as well. To this end, we define a random variable $y_{ij} = u_i^T u_j$. Though the distribution of $y_{ij}$ is not normal, it has the following mean and variance

$$\mu_{y_{ij}} = \mu_{u_i}^T \mu_{u_j}$$
$$\sigma_{y_{ij}}^2 = tr[\Sigma_{u_i} \Sigma_{u_j}] + \mu_{u_i}^T \Sigma_{u_i} \mu_{u_i} + \mu_{u_j}^T \Sigma_{u_j} \mu_{u_j}. \quad (16)$$

Hence, we choose to approximate $y_{ij}$'s distribution with $N_{y_{ij}}(\mu_{y_{ij}}, \sigma_{y_{ij}}^2)$. Then, $-\sigma_{y_{ij}}^2$ can be used as a confidence level of the similarity score.

BSG further enables the applications of other similarity types. For example, we can approximate $p(d_{ij} = 1 | D)$ by approximating the marginalization

$$p(d_{ij} = 1 | D) = \int p(d_{ij} = 1, u_i, u_j | D) du_i du_j =$$
$$\int p(d_{ij} = 1 | u_i, u_j) p(u_i | D) p(u_j | D) du_i du_j \approx$$
$$\int \sigma(u_i^T u_j) q(u_i) q(u_j) du_i du_j \approx \int \sigma(y_{ij}) p(y_{ij}) dy_{ij} \approx \quad (17)$$
$$\sigma\left(\mu_{y_{ij}} / \sqrt{1 + \sigma_{y_{ij}}^2 \pi / 8}\right).$$

where $\mu_{y_{ij}}$ and $\sigma_{y_{ij}}^2$ are given by Eq. (16) and the last three approximations follow from the VB approximation, Eq. (16) and (MacKay 1992), respectively.

Another option is to apply a similarity measure that is based on a symmetric version of the KL divergence between two multivariate normal distributions

## BSG Algorithm

*Input:*

- $m$ — target representation dimension
- $T$ — input text, given as sequence of sequences of words
- $\tau$ — prior precision
- $K$ — maximum number of iterations
- $c_{max}$ — maximal window size
- $l$ — number of most frequent words to be considered in the vocabulary
- $\rho$ — subsampling parameter
- $N$ — negative to positive ratio
- $\kappa$ — number of iterations to apply without performing stochastic updates (in the beginning)
- $\varepsilon$ — stopping threshold
- $\gamma$ — decay parameter

*Output:*

$Q = \{\mu_{u_i}, \mu_{v_i}, \Sigma_{u_i}, \Sigma_{v_i}\}_{i=1}^{l}$ - parameters of the distributions $Q = \{q_{u_i}, q_{v_i}\}_{i=1}^{l}$

**1.** Create a set $W = \{w_i\}_{i=1}^{l}$ of the $l$ most frequent words in $T$ and discard all other words from $T$.

**2. for** $i \leftarrow 1$ **to** $l$

  **2.1** $\mu_{u_i} \sim N(0, I)$, $\mu_{v_i} \sim N(0, I)$, $P_{u_i} \leftarrow I$, $P_{v_i} \leftarrow I$

**3.** Compute $p_{uni}^{3/4}$ over $W$ using $T$ as described in Section 2.1.

**4.** $k \leftarrow 1 - \kappa$, $K \leftarrow K - \kappa$ // *first $\kappa$ iterations are performed without stochastic updates*

**5. repeat**

  **5.1.** $T_{sub} \leftarrow$ Subsample ($T$), as described in Section 2.2

  **5.2. for** $i \leftarrow 1$ **to** $l$

    **5.2.1.** $I_{u_i} \leftarrow \phi, I_{v_i} \leftarrow \phi$

    **5.2.2.** $P_{u_i}^{prev} \leftarrow P_{u_i}$, $P_{v_i}^{prev} \leftarrow P_{v_i}$, $r_{u_i}^{prev} \leftarrow r_{u_i}$, $r_{v_i}^{prev} \leftarrow r_{v_i}$ // *save values from the previous iteration*

  **5.3.** Create $I_P$ based on $T_{sub}$ as described in Section 4.2  // *positive sampling*

  **5.4. for** $(i, j)$ **in** $I_P$

    **5.4.1.** $I_{u_i} \leftarrow I_{u_i} \cup \{j\}$, $I_{v_j} \leftarrow I_{v_j} \cup \{i\}$, $d_{ij} \leftarrow 1$

    **5.4.2. for** $n \leftarrow 1$ **to** $N$ // *negative sampling*

      **5.4.2.1.** Sample a negative word index $z$ according to $p_{uni}^{3/4}(w_z)$ s.t. $(i, z) \notin I_p$

      **5.4.2.2.** $I_{u_i} \leftarrow I_{u_i} \cup \{z\}$, $I_{v_z} \leftarrow I_{v_z} \cup \{i\}$, $d_{iz} \leftarrow -1$

  **5.5. if** $k > 0$ **then** $\beta \leftarrow k^{-\gamma}$ **else** $\beta \leftarrow 1$ // *stochastic updates condition*

  **5.6. parfor** $i \leftarrow 1$ **to** $l$ // *parallel for loop*

    **5.6.1.** Compute $P_{u_i}, r_{u_i}$ using Eq. (14)

    **5.6.2.** $P_{u_i} \leftarrow \beta P_{u_i} + (1 - \beta) P_{u_i}^{prev}$, $r_{u_i} \leftarrow \beta r_{u_i} + (1 - \beta) r_{u_i}^{prev}$

    **5.6.3.** $\Sigma_{u_i} \leftarrow P_{u_i}^{-1}$

    **5.6.4.** $\mu_{u_i} \leftarrow \Sigma_{u_i} r_{u_i}$

    **5.6.5.** $\Sigma_{u_i} \leftarrow diag[diag[\Sigma_{u_i}]]$

  **5.7.** Apply a symmetric version of step 5.6 to $\{q_{v_i}\}_{i=1}^{l}$ parameters

**until** $k > K$ or $\sum_{i=1}^{l} \|r_{u_i} - r_{u_i}^{prev}\|_2 < \varepsilon$ and $\sum_{i=1}^{l} \|r_{v_i} - r_{v_i}^{prev}\|_2 < \varepsilon$

**Figure 1**: The BSG algorithm

$$sim_{symKL}(q_{u_i}, q_{u_j}) = -D_{KL}(q_{u_i} \| q_{u_j}) - D_{KL}(q_{u_j} \| q_{u_i}) \quad (18)$$

where $D_{KL}(q_{u_i} \| q_{u_j})$ has the following closed form solution (Bishop 2006)

$$D_{KL}(q_{u_i} \| q_{u_j}) = \frac{1}{2}\{\log|\Sigma_{u_j}| - \log|\Sigma_{u_i}| +$$
$$(\mu_{u_j} - \mu_{u_i})^T \Sigma_{u_j}^{-1} (\mu_{u_j} - \mu_{u_i})^T + tr[\Sigma_{u_j}^{-1}\Sigma_{u_i}] - m\}.$$

Note that the application of the BSG algorithm to general item similarity tasks is straightforward. The only requirement is that the data is given in the same format. Specifically, every sentence of words in the data is replaced with a sequence of items. Moreover, if the items are given as sets, for each sequence, the window size should be set to the length of the sequence. This results in a Bayesian version of item2vec (Barkan and Koenigstein 2016).

## 5 Experimental Setup and Results

In this section, we compare between the BSG and SG algorithms (for SG we used the word2vec[1] implementation). The algorithms are evaluated on two different tasks: the word similarity task (Finkelstein et al. 2002) and the word analogy task (Pennington, Socher, and Manning 2014).

The word similarity task requires to score pairs of words according to their relatedness. For each pair, a ground truth similarity score is given. The similarity score we used for both models is the cosine similarity. Specifically, for BSG we observed no significant improvement, when applying the similarities from Eqs. (17) and (18), instead of the cosine similarity. In order to compare between the BSG and SG methods, we compute for each method the Spearman (Spearman 1987) rank correlation coefficient with respect to the ground truth.

The word analogy task is essentially a completion task: a bank of questions in the form of ' $w_a$ is to $w_b$ as $w_c$ is to ?' is given, where the task is to replace ? with the correct word $w_d$. The questions are divided to syntactic questions such as 'onion is to onions as lion is to lions' and semantic questions, e.g. 'Berlin is to Germany as London is to England'.

The method we use to answer the questions is by reporting the word $w_d$ that gives the highest cosine similarity score between $u_d$ and $u_? = u_b - u_a + u_c$. For the BSG and SG models we used $\mu_{u_i}$ and $u_i$ as the representation for the word $w_i$, respectively.

[1] https://code.google.com/p/word2vec

### 5.1 Datasets

We trained both models on the corpus from (Chelba et al. 2014). In order to accelerate the training process, we limited our vocabulary to the 30K most frequent words in the corpus and discarded all other words. Then, we randomly sampled a subset of 2.8M 'sentences' that results in a total text length of 66M words.

The word similarity evaluation includes several different datasets: WordSim353 (Finkelstein et al. 2002), SCWS (Huang et al. 2012), Rare Words (Luong, Socher, and Manning 2013), MEN (Bruni, Tran, and Baroni 2014) and SimLex999 (Hill, Reichart, and Korhonen 2015). The reader is referred to the references for further details about these datasets. For each combination of dataset and method, we report the Spearman rank correlation (x100).

The word analogy evaluation dataset (Mikolov et al. 2013) consists of 14 distinct groups of analogy questions, where each group contains a different number of questions. Both models were evaluated on an effective set that contains 14122 questions (all questions that contain out-of-vocabulary words were discarded).

### 5.2 Parameter Configuration

The same parameters configuration was set for both systems. Specifically, we set the target representation dimension $m = 40$, maximal window size $c_{\max} = 4$, subsampling parameter $\rho = 10^{-5}$, vocabulary size $l = 30000$ and negative to positive ratio $N = 1$. For BSG, we further set $\tau = 1$, $\kappa = 10$ and $\gamma = 0.7$ (note that BSG is quite robust to the choice of $\gamma$ as long as $0.5 < \gamma \leq 1$). Both models were trained for $K = 40$ iterations (we verified their convergence after ~30 iterations). In order to mitigate the effect of noise in the results, we trained 10 different instances of BSG and SG and report the average score that is obtained for each entry in the tables.

### 5.3 Results

Table 1 presents the (average) Spearman rank correlation score (x100) obtained by BSG and SG on the word similarity task for various datasets. Table 2 presents the (average) percentage of correct answers for each model per questions group on the word analogy task. We see that the models are competitive where BSG results in a better total accuracy. Examining the results from both tables, we notice that in most cases, BSG achieves better results than SG. This might be explained by the fact that BSG leverages information from second moments as well.

Comparing our results with the literature (Pennington, Socher, and Manning 2014; Mikolov et al. 2013), we see that the scores obtained by both models are lower. This might be explained by several reasons:

TABLE 1: A COMPARISON BETWEEN BSG AND SG ON VARIOUS WORD SIMILARITY DATASETS

| Method | WordSim353 (Finkelstein et al. 2002) | SCWS (Huang et al. 2012) | Rare Words (Luong, Socher, and Manning 2013) | MEN (Bruni, Tran, and Baroni 2014) | SimLex999 (Hill, Reichart, and Korhonen 2015) |
|---|---|---|---|---|---|
| SG | 58.6 | **59.4** | 51.2 | 60.9 | 26.4 |
| BSG | **61.1** | 59.3 | **52.5** | **61.1** | **27.3** |

TABLE 2: A COMPARISON BETWEEN BSG AND SG ON THE WORD ANALOGY TASK

| Questions group name | BSG (%) | SG (%) |
|---|---|---|
| Capital common countries | **62.4** | 59.5 |
| Capital world | **53.2** | 50.2 |
| Currency | **7.1** | 3.6 |
| City in state | 14.3 | **17.9** |
| Family | 55.7 | **63.3** |
| Adjective to adverb | **20.1** | 15.2 |
| Opposite | 6.7 | 6.7 |
| Comparable | **47.2** | 43 |
| Superlative | 41 | **48.1** |
| Present participle | **39** | 36.2 |
| Nationality adjective | **88.9** | 82.3 |
| Past tense | **43.7** | 39.2 |
| Plural | **46.9** | 42.3 |
| Plural verbs | **29.4** | 29.1 |
| **Total** | **45.1** | 42.5 |

First, we use a smaller corpus of 66M words vs. 1-30B in (Pennington, Socher, and Manning 2014; Mikolov et al. 2013). Secondly, the target representation dimension we use is 40 vs. 100-600 in (Pennington, Socher, and Manning 2014; Mikolov et al. 2013). Therefore, we believe that the performance of our models can be improved significantly by increasing the representation dimension as well as the amount of training data. Recall that our main goal is to show that BSG is an effective word embedding method and provides competitive results when compared to the SG method.

## 6 Conclusion

In this paper, we introduced the BSG algorithm that is based on a VB solution to the SG objective. We provide the mathematical derivation of the proposed solution as well as step by step algorithm that is straightforward to implement. Furthermore, we propose several density based similarities measures. We demonstrate the application of BSG on various linguistic datasets and present experimental results that show BSG and SG are competitive.

## References

Barkan, Oren, Yael Brumer, and Noam Koenigstein. 2016. "Modelling Session Activity with Neural Embedding." In *RecSys Posters*.

Barkan, Oren, and Noam Koenigstein. 2016. "Item2Vec: Neural Item Embedding for Collaborative Filtering." Article. *arXiv Preprint arXiv:1603.04259*.

Bishop, Christopher M. 2006. *Pattern Recognition and Machine Learning*. Pattern Recognition. Vol. 4. doi:10.1117/1.2819119.

Bruni, Elia, Nam Khanh Tran, and Marco Baroni. 2014. "Multimodal Distributional Semantics." *Journal of Artificial Intelligence Research* 49: 1–47. doi:10.1613/jair.4135.

Chelba, Ciprian, Tomas Mikolov, Mike Schuster, Qi Ge, Thorsten Brants, Phillipp Koehn, and Tony Robinson. 2014. "One Billion Word Benchmark for Measuring Progress in Statistical Language Modeling." In *Proceedings of the Annual Conference of the International Speech Communication Association, INTERSPEECH*, 2635–39.

Collobert, Ronan, and Jason Weston. 2008. "A Unified Architecture for Natural Language Processing: Deep Neural Networks with Multitask Learning." In *Proceedings of the 25th International Conference on Machine Learning*, 160–67. doi:10.1145/1390156.1390177.

Finkelstein, Lev, Evgeniy Gabrilovich, Yossi Matias, Ehud Rivlin, Zach Solan, Gadi Wolfman, and Eytan Ruppin. 2002. "Placing Search in Context: The Concept Revisited." *ACM Transactions on Information Systems* 20 (1): 116–31. doi:10.1145/503104.503110.

Frome, Andrea, Gs Corrado, and Jonathon Shlens. 2013. "Devise: A Deep Visual-Semantic Embedding Model." *Advances in Neural Information Processing Systems*, 1–11.

Garten, Justin, Kenji Sagae, Volkan Ustun, and Morteza Dehghani. 2015. "Combining Distributed Vector Representations for Words." Inproceedings. In *Proceedings of NAACL-HLT*, 95–101.

Hill, Felix, Roi Reichart, and Anna Korhonen. 2015. "SimLex-999: Evaluating Semantic Models with (Genuine) Similarity Estimation." *Computational Linguistics* 41 (4): 665–95. doi:10.1162/COLI.

Huang, Eric H, Richard Socher, Christopher D Manning, and Andrew Ng. 2012. "Improving Word Representations via Global

Context and Multiple Word Prototypes." *Proceedings of the 50th Annual Meeting of the Association for Computational Linguistics: Long Papers-Volume 1*, 873–82.

Jaakkola, Tommi S, and Michael I Jordan. 1996. "A Variational Approach to Bayesian Logistic Regression Models and Their Extensions." *Aistats*, no. AUGUST 2001. doi:10.1.1.49.5049.

Lazaridou, Angeliki, The Pham Nghia, and Marco Baroni. 2015. "Combining Language and Vision with a Multimodal Skip-Gram Model." *Proceedings of Human Language Technologies: The 2015 Annual Conference of the North American Chapter of the ACL, Denver, Colorado, May 31 – June 5, 2015*, 153–63.

Luong, Minh-Thang, Richard Socher, and Christopher D. Manning. 2013. "Better Word Representations with Recursive Neural Networks for Morphology." *CoNLL-2013*, 104–13.

MacKay, David J. C. 1992. "The Evidence Framework Applied to Classification Networks." *Neural Computation* 4 (5): 720–36. doi:10.1162/neco.1992.4.5.720.

Mikolov, Tomas, Kai Chen, Greg Corrado, and Jeffrey Dean. 2013. "Distributed Representations of Words and Phrases and Their Compositionality." *Nips*, 1–9. doi:10.1162/jmlr.2003.3.4-5.951.

Mnih, Andriy, and Geoffrey E. Hinton. 2008. "A Scalable Hierarchical Distributed Language Model." *Advances in Neural Information Processing Systems*, 1–8.

Paquet, Ulrich, and Noam Koenigstein. 2013. "One-Class Collaborative Filtering with Random Graphs." *Proceedings of the 22nd International Conference on World Wide Web*, 999–1008.

Pennington, Jeffrey, Richard Socher, and Christopher D Manning. 2014. "GloVe: Global Vectors for Word Representation." *Proceedings of the 2014 Conference on Empirical Methods in Natural Language Processing*, 1532–43. doi:10.3115/v1/D14-1162.

Robbins, Herbert, and Sutton Monro. 1951. "A Stochastic Approximation Method." *The Annals of Mathematical Statistics* 22 (3): 400–407. doi:10.1214/aoms/1177729586.

Spearman, C. 1987. "The Proof and Measurement of Association between Two Things. By C. Spearman, 1904." *The American Journal of Psychology* 100 (3–4): 441–71. doi:10.1037/h0065390.

Vilnis, Luke, and Andrew Mccallum. 2015. "Word Represenation via Guassian Embedding." In *In Proceedings of the ICLR 2015*, 1–12.

Zhang, J, J Salwen, M Glass, and A Gliozzo. 2014. "Word Semantic Representations Using Bayesian Probabilistic Tensor Factorization." In *Proceedings of the 2014 Conference on Empirical Methods in Natural Language Processing (EMNLP)*.